\documentclass[11pt,a4paper]{article}
\usepackage[hyperref]{acl2021}
\usepackage{times}
\usepackage{latexsym}

\usepackage{times}
\usepackage{latexsym}
\usepackage{microtype}
\usepackage{amsfonts, amsmath, amsthm}
\usepackage{breqn}
\usepackage{makecell}
\usepackage{scalerel}
\usepackage{subfig}

\usepackage{xspace}
\usepackage{url}
\usepackage{xfrac}
\usepackage{booktabs}
\usepackage{tikz}
\newsavebox{\tempbox}
\usepackage{arydshln}
\usepackage{enumerate}
\usepackage[inline]{enumitem}
\usepackage{float}
\usepackage{adjustbox}
\usepackage{microtype}

\definecolor{darkgreen}{RGB}{0,160,0}
\definecolor{darkred}{RGB}{160,0,0}
\newcommand{\dd}[1]{\textcolor{darkgreen}{\bf\fontsize{7.5}{12} \selectfont \! \!$\pm$ #1}}
\newcommand{\dm}[1]{\textcolor{darkred}{\bf\fontsize{7}{12} \selectfont \! \!- \!#1}}
\newcommand{\dpl}[1]{\textcolor{darkgreen}{\bf\fontsize{7}{12} \selectfont \! \!$+$\! #1}}

\aclfinalcopy % Uncomment this line for the final submission
\def\aclpaperid{1992} %  Enter the acl Paper ID here

\newcommand{\calY}{{\cal Y}}

\newcommand{\xx}{\mathbf{x}}
\newcommand{\zz}{\mathbf{z}}
\newcommand{\bg}{\mathbf{g}}
\newcommand{\embedxx}{X^{(e)}}

\newcommand{\yhat}{\hat{y}}

\newcommand{\defeq}{\mathrel{:\mkern-0.25mu=}}
\newcommand{\ent}{\mathrm{H}}
\newcommand{\ra}[1]{\renewcommand{\arraystretch}{#1}}

\newcommand{\cc}{\mathbf{c}}
\newcommand{\hh}{\mathbf{h}}

\newcommand{\qq}{\mathbf{q}}
\newcommand{\pp}{\mathbf{p}}

\newcommand{\R}{\mathbb{R}}
\newcommand{\va}{\mathbf{a}}
\newcommand{\vv}{\mathbf{v}}
\newcommand{\weights}{{\boldsymbol \alpha}}

\newcommand{\vocab}{\mathcal{V}}
\newcommand{\softphi}{\phi_{\mathrm{soft}}}
\newcommand{\sparsephi}{\phi_{\mathrm{sparse}}}
\newcommand{\sparsegenphi}{\phi_{\mathrm{sparseg}}}

\newcommand{\defn}[1]{\textbf{#1}}
\newcommand{\norm}[1]{\left\lVert#1\right\rVert}

\DeclareMathOperator*{\argmin}{argmin}

\usepackage{cleveref}
\crefname{section}{\S}{\S\S}
\Crefname{section}{\S}{\S\S}
\crefname{table}{Table}{}
\crefname{figure}{Fig.}{}
\crefname{algorithm}{Algorithm}{}
\crefname{equation}{Eq.}{Eq.}
\crefname{appendix}{App.}{}
\crefname{prop}{Proposition}{}
\crefformat{section}{\S#2#1#3}  % remove space between section symbol and the number

\usepackage[disable]{todonotes}
\makeatletter
\newcommand*\iftodonotes{\if@todonotes@disabled\expandafter\@secondoftwo\else\expandafter\@firstoftwo\fi}  % defines \iftodonotes{<true>}{<false>}, thanks to https://tex.stackexchange.com/questions/126559/conditional-based-on-packageoption
\makeatother

\newcommand{\note}[4][]{\todo[author=#2,color=#3,size=\scriptsize,fancyline,caption={},#1]{#4}} % default note settings, used by macros below.
\newcommand{\ryan}[2][]{\note[#1]{ryan}{violet!40}{#2}}
\newcommand{\clara}[2][]{\note[#1]{clara}{orange}{#2}}

\everypar{\looseness=-1}

\usepackage{marvosym}
\usepackage{amssymb}
\newcommand{\ucambridge}{\bigstar}
\newcommand{\ethz}{\text{\Lightning}}
\newcommand{\ucop}{\text{\Moon}}

\title{Is Sparse Attention more Interpretable?}

\author{Clara Meister$^{\ethz}$ Stefan Lazov$^{\ucambridge}$ Isabelle Augenstein$^{\ucop}$  Ryan Cotterell$^{\ucambridge,\ethz}$ \\
  $^{\ethz}$ETH Z\"{u}rich~\;~ $^{\ucambridge}$University of Cambridge~\;~$^{\ucop}$University of Copenhagen%
   \\
  \texttt{meistecl@inf.ethz.ch},~\;~ \texttt{stefan.lazov@cantab.net},~\;~\\ \texttt{augenstein@di.ku.dk},~\;~ \texttt{ryan.cotterell@inf.ethz.ch}
}

\date{}

\begin{document}
\maketitle
\begin{abstract}
    Sparse attention has been claimed to increase model interpretability under the assumption that it highlights influential inputs. Yet the attention distribution is typically over representations internal to the model rather than the inputs themselves, suggesting this assumption may not have merit. We build on the recent work exploring the interpretability of attention; we design a set of experiments to help us understand how sparsity affects our ability to use attention as an explainability tool. On three text classification tasks, we verify that only a weak relationship between inputs and co-indexed intermediate representations exists---under sparse attention and otherwise. Further, we do not find any plausible mappings from sparse attention distributions to a sparse set of influential inputs through other avenues. Rather, we observe in this setting that inducing sparsity may make it less plausible that attention can be used as a tool for understanding model behavior.\looseness=-1
\end{abstract}
\section{Introduction}
Interpretability research in natural language processing (NLP) is becoming increasingly important as complex models are applied to more and more downstream decision making tasks.
In light of this, many researchers have turned to the attention mechanism, which has not only led to impressive performance improvements in neural models, but has also been claimed to offer insights into how models make decisions.
Specifically, a number of works imply or directly state that one may inspect the attention distribution to determine the amount of influence each input token has in a model's decision-making process \cite[][\emph{inter alia}]{xie-etal-2017-interpretable, mullenbach-etal-2018-explainable,niculae2018}.\looseness=-1

\begin{figure}
\centering
    \includegraphics[width=\linewidth]{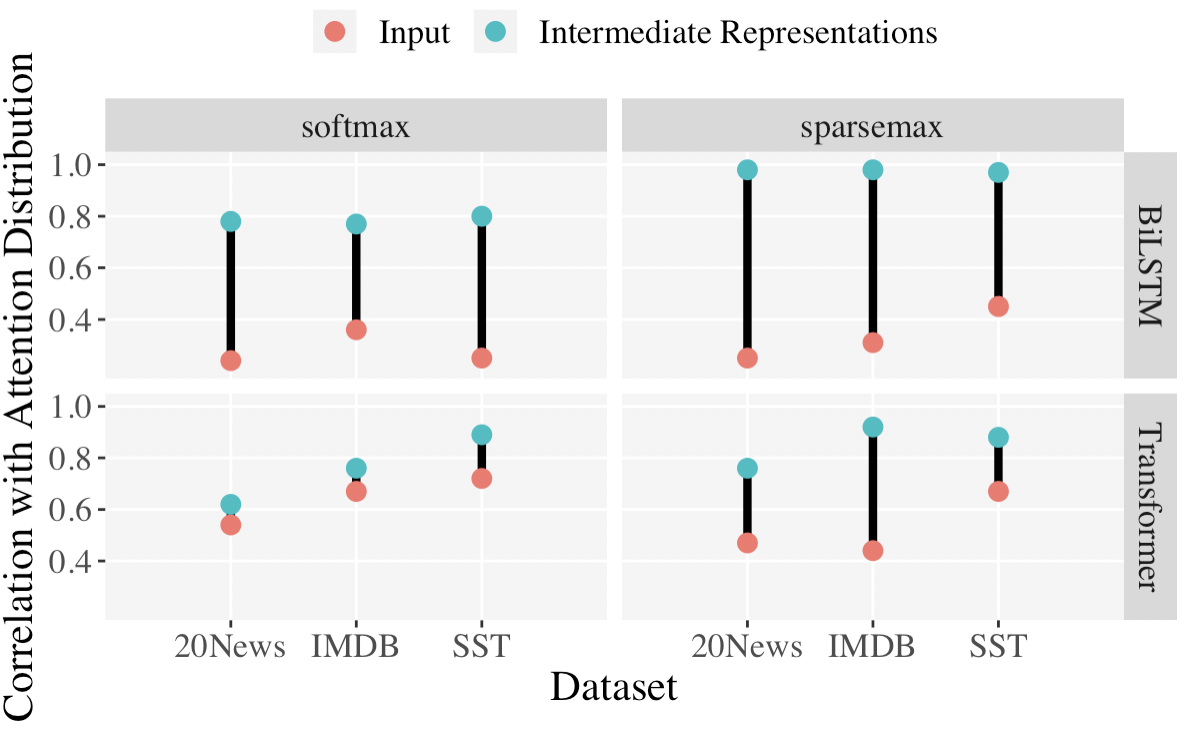}
  \caption{Correlation between the attention distribution and gradient-based FI measures. We see a notably stronger correlation between attention and FI of intermediate representation than of inputs across all models.}
    \label{fig:correlation}
    \vspace{0.5em}
\end{figure}

Many lines of work have gone on to exploit this assumption when building their own ``interpretable'' models or analysis tools \cite{yang-etal-2016-hierarchical, tu-etal-2016-modeling, gender_bias}; one subset has even tried to make models with attention \textit{more} interpretable  by inducing sparsity---a common attribute of interpretable models \cite{lipton_interp, Rudin2019}---in attention weights, with the motivation that this allows model decisions to be mapped to a \emph{limited} number of items \cite{pmlr-v48-martins16,malaviya-etal-2018-sparse,ieee_sparsity}. 
Yet, there lacks concrete reasoning or evidence that sparse attention weights leads to more interpretable models: customarily, attention is not directly over the model's inputs, but rather over some representation \emph{internal} to the model, e.g. the hidden states of a recurrent network or contextual embeddings of a Transformer (see \cref{fig:correlation}). Importantly, these internal representations do not solely encode information from the input token they are co-indexed with \cite{salehinejad2017recent,brunner2019identifiability}, but rather from a range of inputs. This presents the question: if internal representations themselves may not be interpretable, can we actually deduce anything from ``interpretable'' attention weights?\looseness=-1

We build on the recent line of work challenging the validity of attention-as-explanation methods \cite[][\textit{inter alia}]{jain2019attention,serrano-smith-2019-attention,grimsley-etal-2020-attention} and specifically examine how \emph{sparsity} affects their observations. 
To this end, we introduce a novel entropy-based metric to measure the \emph{dispersion} of inputs' influence, rather than just their magnitudes. Through experiments on three text classification tasks, utilizing both LSTM and Transformer-based models, we observe how sparse attention affects the results of \citet{jain2019attention} and \citet{wiegreffe2019attention}, additionally exploring whether it allows us to identify a core set of inputs that are important to models' decisions. 
We find we are unable to identify such a set when using sparse attention; rather, it appears that encouraging sparsity may simultaneously encourage a higher degree of contextualization in intermediate representations.
We further observe a decrease in the correlation between the attention distribution and input feature importance measures, which exacerbates issues found by prior works. The primary conclusion of our work
is that we should not believe sparse attention enhances model interpretability until we have concrete reasons to believe so; in this preliminary analysis, we do not find any such evidence.\looseness=-1

\section{Attention-based Neural Networks}\label{sec:attn-nn}
We consider inputs $\xx = x_1 \cdots x_n \in \vocab^n$ of length $n$ where the tokens from taken from an alphabet $\vocab$. 
We denote the embedding of $\xx$, e.g., its one hot encoding or (more commonly) a linear transformation of its one-hot encoding with an embedding matrix $E \in \R^{ d \times |\vocab|}$, as $\embedxx \in \R^{d \times n}$. Our embedded input $\embedxx$ is then fed to an encoder, which produces $n$ intermediate representations $I = [\hh_1; \dots; \hh_n] \in \R^{m \times n}$, where $\hh_i \in \R^m$ and $m$ is a hyperparameter of the encoder. This transformation is quite architecture dependent. 

An alignment function $A(\cdot, \cdot)$ maps a \defn{query} $\qq$ and a \defn{key} $K$ to weights $\va^{(t)}$ for a decoding time step $t$; we subsequently drop $t$ for simplicity.\ryan{I added this because $t$ magically disappeared. Is that okay?}
In colloquial terms, $A$ chooses which values of $K $ should receive the most attention based on $\qq$, which is then represented in the vector $\va^{(t)} \in \R^n$. For the NLP tasks we consider, we have $K = I = [\hh_1; \dots; \hh_n]$, the encoder outputs. A query $\qq$ may be, e.g., a representation of the question in question answering.\looseness=-1

The weights $\va$ are %then
projected to sum to 1, which results in the  \defn{attention distribution} $\weights$. Mathematically, this is done via a projection onto the probability simplex using a projection function $\phi$, e.g., softmax or sparsemax. We then compute the \defn{context vector} as $\cc = \sum_{i=1}^n \alpha_i\,\hh_i$. 
This context vector is fed to a decoder, whose structure is again architecture dependent, which generates a (possibly unnormalized) probability distribution over the set of labels $\calY$, where $\calY$ is defined by the task.\looseness=-1 

\paragraph{Attention.}
We experiment with two methods of constructing an attention distribution: (1) additive attention, proposed by \newcite{bahdanau2014neural}: $A(K, \qq)_i = \vv^{\top} \tanh(W_1 K_i + W_2 \qq)$ and (2) the scaled dot product alignment function,  as in the Transformer network: $A(K, \qq) = \frac{K^{\top} \qq}{\sqrt{m}}$ where  $\vv \in \R^{l}$ and $W_1, W_2 \in \R^{l\times m}$ are weight matrices. 
Note that the original (without attention) neural encoder--decoder architecture, as in \newcite{sutskever_rnn}, can be recovered with alignment function $A(\cdot, \cdot) = [0, \dots, 0,1]$, i.e., only the last of the $n$ intermediate representations is given to the decoder.
\paragraph{Projection Functions.}
A projection function $\phi$ takes the output of the alignment function and maps it to a valid probability distribution: $\phi: \R^n \rightarrow \Delta^{n-1}$. The standard projection function is softmax:
\begin{align}
    \softphi(\zz) &= \frac{\exp(\zz)}{\sum_{i \in [n]}\exp(z_i)}\\
    &= \argmin_{\pp \in \Delta^{n-1}} \left(  
    \sum_{i \in [n]} p_i\log p_i - \pp^{\top} \zz  \right) \nonumber
\end{align}
However, softmax leads to non-sparse solutions as an entry $\softphi(\zz)_i$ can only be $0$ if $x_i\!=\!-\infty$. Alternatively, \newcite{pmlr-v48-martins16} introduce \defn{sparsemax}, which can output sparse distributions:
\begin{equation}\label{eq:spmax}
    \sparsephi(\zz) = \argmin_{\pp\in \Delta^{n-1}} \norm{\pp- \zz }^2_2
\end{equation}
In words, sparsemax directly maps $\zz$ onto the probability simplex, which often leads to solutions on the boundary, i.e. where at least one entry of $\mathbf{p}$ is $0$. The formulation of sparsemax in \cref{eq:spmax} does not give us an explicit medium for controlling the \emph{degree} of sparsity. The $\alpha$-entmax \cite{peters-etal-2019-sparse} and \defn{sparsegen} \cite{sparsegen} transformations fill this gap; we employ the latter:\looseness=-1
\begin{equation}
    \sparsegenphi(\zz) = \argmin_{\mathbf{p} \in \Delta^{n-1}} \norm{\mathbf{p}- g(\zz) }^2_2 - \lambda\norm{\mathbf{p}}^2_2
\end{equation}
\noindent where the degree of sparsity can be tuned via the hyperparameter $\lambda \in (-\infty, 1)$. Note that a larger $\lambda$ encourages more sparsity in the minimizing solution.\looseness=-1

\section{Model Interpretability}
Model interpretability and explainability have been framed in different ways \cite{CSI19}---as model understanding tasks, where (spurious) features learned by a model are identified, or as decision understanding tasks, where explanations for particular instances are produced. We consider the latter in this paper.  Such tasks can be framed as generative, where models generate free text explanations \cite{camburu2018snli,kotonya-toni-2020-explainable,atanasova-etal-2020-generating}, 
or as post-hoc interpretability methods, where salient portions of the input are highlighted \cite{lipton_interp,deyoung-etal-2020-eraser,atanasova-etal-2020-diagnostic}.

As there does not exist a clearly superior choice for framing decision understanding for NLP tasks %quantifying interpretability in NLP tasks 
\cite{MILLER20191,carton-etal-2020-evaluating,jacovi-goldberg-2020-towards}, we follow a substantial body of prior work in considering the post-hoc definition of interpretability based on local methods proposed by \newcite{lipton_interp}. This definition is naturally operationalized through feature importance metrics and meta models \cite{jacovi-goldberg-2020-towards}.
Further, we acknowledge the specific requirement that an interpretable model obeys some set of structural constraints of the domain in which it is used, such as monotonicity or physical constraints \cite{Rudin2019}. For NLP tasks such as sentiment analysis or topic classification, such constraints may logically include the utilization of \emph{only} a few key words in the input when making a decision, in which case, knowing the magnitude of the influence each input token has on a model's prediction through, e.g., feature importance metrics, may suffice to verify the model obeys such constraints. 
While this collective definition is limited \cite{doshi_velez_interpretable,10.1145/3236009,Rudin2019}, we posit that if attention cannot provide model interpretability at this level, then it would likewise not be able to under more rigorous constraints.\looseness=-1

\subsection{Measures of Feature Importance}
\paragraph{Gradient-Based.}
Gradient-based measures of feature importance  \cite[F1;][]{ DBLP:journals/jmlr/BaehrensSHKHM10,DBLP:journals/corr/SimonyanVZ13,poerner-etal-2018-evaluating} use the gradient of a function's output w.r.t. a feature to measure the importance of that feature. %\footnote{If we were analyzing whether the input contributed positively or negatively to a given 
In the case of an attentional neural network for binary classification $f(\cdot)$, we can take the gradient of $f$ w.r.t. the variable $\xx$ and evaluate at a specific input $\xx'$ to gain a sense of how much influence each $x_i'$ had on the outcome $\yhat = f(\xx')$. These measures are not restricted to the relationship between inputs $x_i$ and the outcome $f(\xx)$; they can also be adapted to measure for effects from and to intermediate representations $\hh_p$.
Formally, our measures are as follows:\looseness=-1
\begin{align}
    \bg_\yhat(x_i) &= \frac{\norm{\frac{\partial f}{\partial \embedxx_i}}_2}{\sum_{k=1}^n\norm{\frac{\partial f}{\partial \embedxx_k}}_2} 
    \label{eq:gradient} \\
    \bg_{\hh_p}(x_i) &= \frac{\norm{\frac{\partial ||\hh_p||_2}{\partial \embedxx_i}}_2}{\sum_{k=1}^n\norm{\frac{ \partial ||\hh_p||_2}{\partial \embedxx_k}}_2}\label{eq:gradient_intermediate}
\end{align}
\noindent where $\bg_\yhat(x_i) \in [0,1]$ and $\bg_{x_i}(\hh_p) \in [0,1]$ represents the gradient-based FI of token $x_i$ on $\yhat$ and intermediate representation $\hh_p$, respectively. Gradient-based methods are often used in explainability techniques, as they have exhibited higher correlation with human judgement than others \cite{atanasova-etal-2020-diagnostic}. Note that we take gradients w.r.t. the embedding of token $x_i$ and that in the latter metric, we measure the influence of $x_i$ on the magnitude of $\hh_p$---a decision we discuss in \cref{app:feature_importance}.\looseness=-1

\paragraph{Erasure-based.}
As a secondary FI metric, we observe how model predictions change when a specific input token is removed (i.e., Leave-One-Out; LOO). For token $x_i$, this can be calculated as:
\begin{equation}
    D_\yhat(x_i) = \frac{|\yhat-\yhat_{-i}|}{\sum_{k=1}^{n}|\yhat-\yhat_{-k}|}
    \label{eq:LOO}
\end{equation}\ryan{Is this denominator right?}\clara{I believe so but someone else should check now :/}
\noindent where $\yhat_{-i}$ is the prediction of a model with input $x_i$ removed. The formula can also be used for intermediate representations; we denote this as $D_\yhat(\hh_i)$.\looseness=-1

\section{Experiments}\label{sec:experiments}

\paragraph{Setup.} We run experiments across several model architectures, attention mechanisms, and datasets in order to understand the effects of induced attentional sparsity on model interpretability.
We use three binary classification datasets: ImDB and SST (sentiment analysis) and 20News (topic classification).  
We use the dataset versions provided by \newcite{jain2019attention}, exactly following their pre-processing steps.  We show a subset of representative results here, with additional results in \cref{app:additional}. Further details, including model architecture descriptions, dataset statistics and baselines accuracies may be found in \cref{sec:setup}.\looseness=-1
\paragraph{Inputs and Intermediate Representations are not Interchangeable.}

We first explore how strongly-related inputs are to their co-indexed intermediate representations. 
A strong relationship on its own may validate the use of sparse attention, as the ability to identify a subset of influential intermediate representations would then directly translate to a set of influential inputs. 
Previous works show that the ``contribution'' of a token $x_i$ to its intermediate representation $\hh_i$ is often quite low for various model architectures \cite{salehinejad2017recent, rnnvis, brunner2019identifiability,tutek-snajder-2020-staying}.
In the context of attention, we find this property to be evinced by the adversarial experiments of \newcite{wiegreffe2019attention} (\textsection 4) and 
\newcite{jain2019attention} (\textsection 4), which we verify in \cref{app:additional}.  They construct adversarial attention distributions by optimizing for divergence from a baseline model's attention distribution by: (1) adopting all of the baseline model's parameters and directly optimizing for divergence and (2) training an entirely new model and optimizing for divergence as part of the training process. The former method leads to a large drop in performance (accuracy) while the latter does not. If we believe the model must encode the same information to achieve similar accuracy, this discrepancy implies that in the latter method, the model likely ``redistributes'' information across encoder outputs (i.e., intermediate representations $\hh_p$), which would suggest token-level information is not tied to a particular $\hh_p$.\looseness=-1
\begin{table}
\ra{1.2}
  \centering
  \setlength\tabcolsep{4.3pt} % default value: 6pt
  \footnotesize
  \adjustbox{width=\linewidth}{
  \begin{tabular}{ @{}lllll@{} }
  \toprule
      & \multicolumn{1}{c}{\bf IMDb} & \multicolumn{1}{c}{\bf 20-News} & 
      \multicolumn{1}{c}{\bf SST}\\
      & $\tilde\ent(\bg_{\hh_i}(\xx))$   
      & $\tilde\ent(\bg_{\hh_i}(\xx))$   
      & $\tilde\ent(\bg_{\hh_i}(\xx))$   \\
    \hline
    BiLSTM (Softmax)         & 0.71 \dd{0.09}  & 0.75 \dd{0.12}        & 0.93 \dd{0.05}  \\
    BiLSTM (Sparsemax)         & 0.72 \dd{0.10}  & 0.68 \dd{0.12}        & 0.91 \dd{0.07}  \\
    \hline
    Transformer (Softmax)         & 0.76 \dd{0.08}  & 0.48 \dd{0.06}        & 0.73 \dd{0.09}  \\
    Transformer (Sparsemax)         & 0.72 \dd{0.09}  & 0.46 \dd{0.06}        & 0.63 \dd{0.08}  \\
    \bottomrule
  \end{tabular} }
  \caption{ Mean entropy of gradient-based FI of input to intermediate representations. Green numbers are std. deviations. Projection functions are parenthesized.\looseness=-1 }
  \label{tab:coindexed_correlation}
\end{table}

As further verification of high degrees of contextualization in attentional models, we report a novel quantification, offering insights into whether individual intermediate representations can be linked primarily to \emph{any} single input---i.e., perhaps not the co-indexed input; we measure the normalized entropy\footnote{We use Shannon entropy $\tilde\ent(p) \defeq -\sum_{x } p(x)\log p(x)$ normalized (i.e. divided) by the maximum possible entropy of the distribution to control for dimension.} of the gradient-based FI of inputs to intermediate representations $\tilde\ent(\bg_{\hh_p}(\xx))\in [0,1]$\ryan{I think this is also not true. Entropy goes up to $\log n$}\clara{yes but we're using 'normalized' entropy here. I think its ok since we specify in the sentence before and the footnote} to gain a sense of how dispersed influence for intermediate representation is across inputs. A value of $1$ would indicate all inputs are equally influential; a value of $0$ would indicate solely a single input has influence on an intermediate representation. 
Results in \cref{tab:coindexed_correlation} show consistently high entropy in the distribution of the influence of inputs $x_i$ on an intermediate representation $\hh_p$ across all datasets, model architectures, and projection functions, which suggests the relationship between intermediate representations and inputs is far from one-to-one in these tasks.\looseness=-1 
\begin{figure}

\centering
    \includegraphics[width=\linewidth]{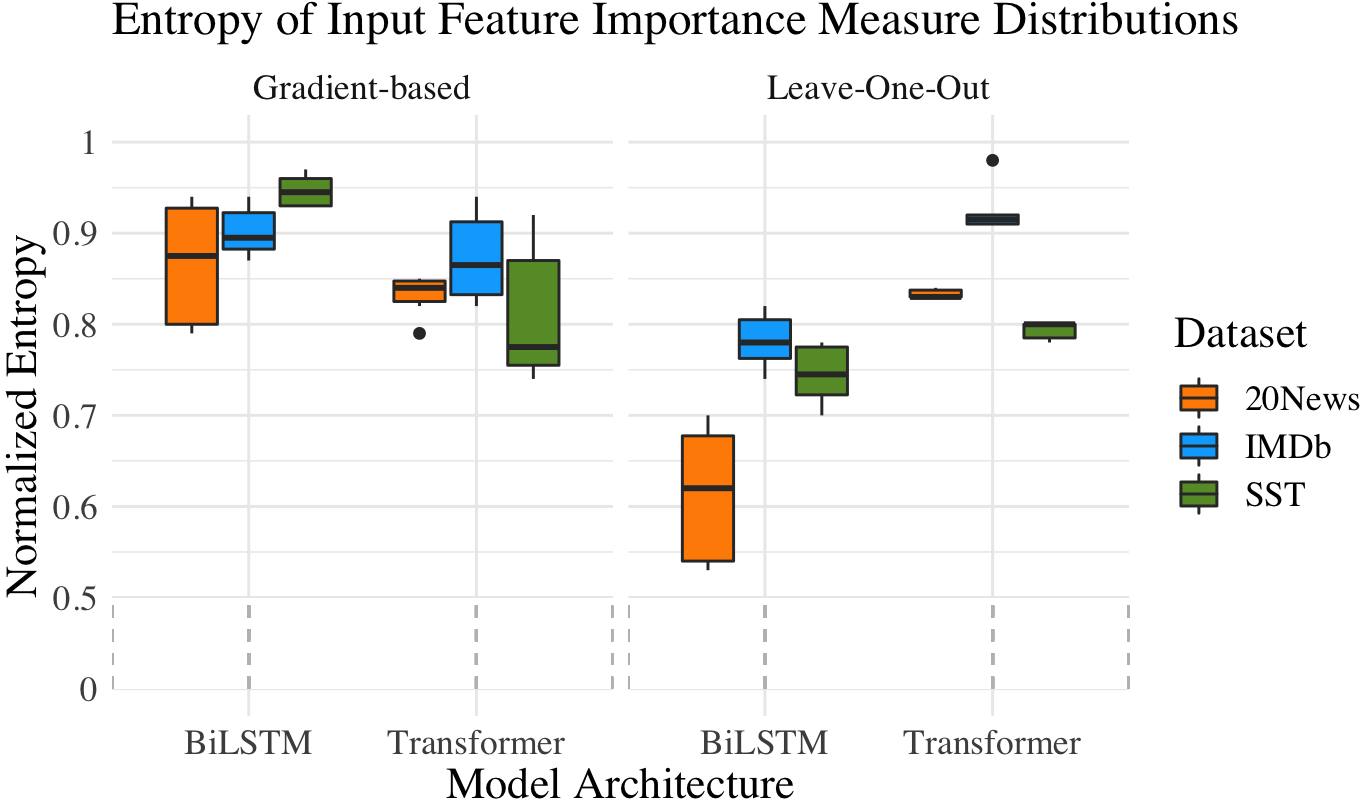}
  \caption{Entropy of gradient-based $\bg_\yhat(\xx)$ and LOO $D_\yhat(\xx)$ FI distributions. %for BiLSTM and Transformer. 
  Results are from models with full spectrum of projection functions.}%; normalized entropy of 1 indicates the uniform distribution and 0 indicates a degenerate distribution (all mass on one point). }
    \label{fig:entropy}
    \vspace{0.5em}
\end{figure}

\paragraph{Sparse Attention $\neq$ Sparse Input Feature Importance.}
Our prior results demonstrated that---even when using sparse attention---we cannot identify a subset of influential inputs directly through intermediate representations; we explore whether a subset can still be identified through FI metrics. In the case where the normalized FI distribution highlights only a few key items, the distribution will, by definition, have low entropy. 
Thus, we explore whether sparse attention leads to lower entropy input FI distributions in comparison to standard attention. We find no such trend;  \cref{fig:entropy} shows that across all models and settings, the entropy of the FI distribution is quite high. Further, we see a consistent \emph{negative} correlation between this entropy and the sparsity parameter of the sparsegen projection (\cref{tab:ent_cor}), implying that entropy of feature importance \emph{increases} as we raise the degree of sparsity in $\weights$.\looseness=-1 
\begin{table}
\ra{1.2}
  \centering
  \setlength\tabcolsep{4.3pt} % default value: 6pt
  \small
  \adjustbox{max width=\linewidth}{
  \begin{tabular}{ @{}llll@{} }
  \toprule
      & \multicolumn{1}{c}{\bf IMDb} & \multicolumn{1}{c}{\bf 20-News} & 
      \multicolumn{1}{c}{\bf SST}\\
    \hline
    BiLSTM (tanh) &-0.935  & -0.675 & -0.866\\
    Transformer (dot)  & -0.830 &  -0.409 &  -0.810  \\
    \bottomrule
  \end{tabular} }
  \caption{ Correlation between sparsegen parameter$^2$ $\lambda$ and entropy of gradient-based input FI $\tilde\ent(\bg_\yhat(\xx))$. \looseness=-1 }
  \label{tab:ent_cor}
\end{table}

\begin{figure}
\centering
    \includegraphics[width=\linewidth]{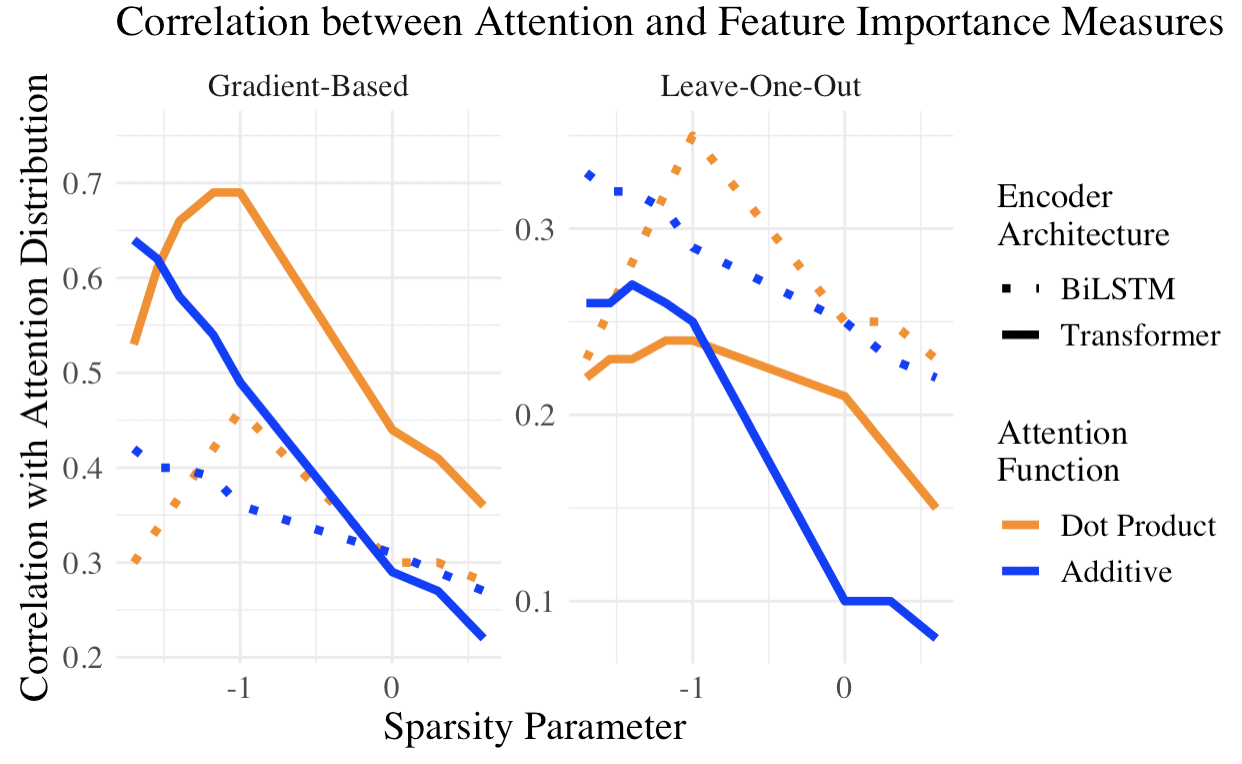}
  \caption{Correlation between the attention distribution and input FI measures as a function of the sparsity penalty $\lambda$ used in the projection function $\sparsegenphi$. $x$-axis is log-scaled for $\lambda < 0$ since $\lambda \in (-\infty, 1)$. Results are from the IMDb dataset. }
    \label{fig:sparsity}
\end{figure}

\paragraph{Correlation between Attention and Feature Importance.}
Finally, we follow the experimental setup of \newcite{jain2019attention}, who postulate that if the attention distribution indicates which inputs influence model behavior, then one may reasonably expect attention to correlate\footnote{We use Kendall's $\tau$-b correlation \cite{10.2307/2282833}.} with FI measures of the input. 
While they find only a weak correlation, we explore how inducing sparsity in the attention distribution affects this result. Surprisingly,
\cref{fig:sparsity} shows a downward trend in this correlation as the sparsity parameter $\lambda$ of the sparsegen projection function is increased. 
As argued by \citet{wiegreffe2019attention}, a lack of this correlation does not indicate attention \emph{cannot} be used as explanation; FI measures are not ground-truth indicators of critical inputs. However, the inverse relationship between input FI and attention is rather surprising. If anything, we may surmise sparsity in $\weights$ leads to \emph{less} faithful explanations from $\weights$.\looseness=-1
From these results, we posit that promoting sparsity in attention distribution may simply lead to the dispersion of information to different intermediate representations, a behavior similar to that seen when constraining attention for divergence from another distribution, i.e., in the adversarial experiments of \citet{wiegreffe2019attention} compared to those of \citet{jain2019attention}. 

\section{Related Work}

The use of attention as an indication of inputs' influence on model decisions may at first seem natural; yet a large body of work has recently challenged this practice. Perhaps the first to do so was \newcite{jain2019attention}, which revealed both a lack of correlation between the attention distribution and well established feature importance metrics and of unique optimal attention weights. \newcite{serrano-smith-2019-attention} contemporaneously found similar results. Subsequently, other studies arrived at similar results: for example, \citet{grimsley-etal-2020-attention} found evidence that causal explanations are not attainable from attention layers over text data; \citet{pruthi-etal-2020-learning} showed that attention masks can be trained to give deceptive explanations. 
We view this work as another such investigation, exploring attention's innate interpretability on a different axis.\looseness=-1

This work also fits into the context of a larger body of interpretability research in NLP, which has challenged the informal use of terms such as faithfulness, plausibility, and explainability \cite[][\emph{inter alia}]{lipton_interp,arrieta:hal-02381211,jacovi2021contrastive} and tried to quantify the reliability of current definitions \cite{atanasova-etal-2020-diagnostic}. While we consider these works in our experimental design---e.g., in our choice of FI metrics---we recognize that further experiments are needed to verify our findings: for example, similar experiments could be performed using the \citet{deyoung-etal-2020-eraser} benchmark for evaluation; other FI metrics, such as selective attention \cite{treviso-martins-2020-explanation} should additionally be considered.\footnote{Notably, \citet{treviso-martins-2020-explanation} found that inducing sparsity in attention \emph{aided} in the usefulness of their metric as a tool for explaining model decisions.}\ryan{Are you missing a period?}
%We leave this for future work. 

\section{Conclusion}
Prior work has cited interpretability as a driving factor for promoting sparsity in attention distributions. 
We explore how induced sparsity affects the ability to use attention as a tool for explaining model decisions. 
In our experiments on text classification tasks, we see that while sparse attention distributions may allow us to pinpoint influential intermediate representations, we are unable to find any plausible mapping from sparse attention to a small, critical set of influential inputs. Rather, we find evidence that inducing sparsity may make it even less plausible to use the attention distribution to interpret model behavior. We conclude that we need further reason to believe sparse attention increases model interpretability as our results do not support such claims.

\section*{Acknowledgements}
We thank the anonymous reviewers for their insightful feedback on the manuscript.
Isabelle Augenstein's research is partially funded by a DFF Sapere Aude research leader grant.
\section*{Ethical Considerations}
Machine learning models are being deployed in an increasing number of sensitive situations. In these settings, it is critical that models are interpretable, so that we can avoid e.g., inadvertent racial or gender bias. Giving a false sense of interpretability can allow models with undesirable (i.e., unethical or unstable) behavior to fly under the radar. We view this work as another critique of interpretability claims and hope our results will encourage the more careful consideration of interpretability assumptions when using machine learning models in practice.\looseness=-1

\bibliography{acl2021,anthology}
\bibliographystyle{acl_natbib}

\appendix
\clearpage
\newpage
\section{Feature Importance Metrics}\label{app:feature_importance}\clara{add in comment on norm here}
Notably, both inputs and intermediate representations are not single variables. Intermediate representations are $m$-dimensional vectors and inputs $\xx$ are embedded as $\embedxx$, meaning each word $x_i$ is represented by a $d$-dimensional vector. Therefore, the gradient of $f$ w.r.t. individual inputs or intermediate representations will likewise be a $d$- (or $m$-) dimensional vector. To come up with a scalar estimate of feature importance, we take the $L_2$-norm of the evaluated gradient.\footnote{Other norms, e.g., the $L_1$-norm, would also be appropriate---we leave the exploration of these to future work.} Subsequently, we normalize over all $x_i$ (or $\hh_i)$ to calculate \emph{relative} feature importance of individual $x_i$ (or $\hh_i)$. The discussed transformation can be mathematically formalized by \cref{eq:gradient_intermediate,eq:gradient}. For intermediate representations, this computation measures the  influence on the magnitude of $\hh_p$ rather than on $\hh_p$ itself. However, we also experimented with measuring the influence directly on each facet of $\hh_p$, taking the magnitude of this vector. We found empirically that the two measures returned nearly identical  results while measuring influence on magnitude was significantly more computationally efficient.

\section{Experimental Setup}\label{sec:setup}
We use exact datasets provided by and based our experimental framework on that of \citet{jain2019attention}, which can be found at \url{https://github.com/successar/AttentionExplanation}. For both comparison and reproducibility, we exactly follow their preprocessing steps, which are described in their paper. Source code, model statistics, and links to datasets can be found at the above link. In the experiments we use a Bidirectional LSTM encoder or a Transformer encoder which has 2 layers with 1 attention head. All hidden dimensions are set to $128$. The models and the training procedure have been implemented by using the PyTorch library \citet{NEURIPS2019_9015}. For training we use the Adam optimizer \citet{adam} with the amsgrad \citet{amsgrad} option enabled. Some important hyperparameters are listed in \cref{tab:hyperparams}; minor tuning was performed in order to reach comparable performance with respect to \citet{jain2019attention} and \citet{wiegreffe2019attention}. An important note regarding this table is that the listed learning rate and weight decay correspond to all model parameters except the ones specifically for the attention mechanism. The latter we train without a weight decay and with either the same or $10$x larger learning rate. 
\begin{table}[!h]
\ra{1.2}
  \centering
  \footnotesize
  \adjustbox{width=\linewidth}{
  \begin{tabular}{ @{}lllll@{} }
  \toprule
      & Train size & Test size & Accuracy (T) & Accuracy (B)\\
    \hline
    IMDb & 25000& 4356 & 0.89& 0.90\\
    20News  & 1426 &334 & 0.91& 0.91\\
    SST  & 6355 &1725 & 0.79& 0.82\\
    \bottomrule
  \end{tabular} }
  \caption{ Dataset statistics and baseline accuracy scores on test sets for Transformer with dot product attention (T) and BiLSTM with additive attention (B). All datasets are in english. }
  \label{tab:stats}
\end{table}

\begin{table}[!h]
\ra{1.2}
  \centering
  \footnotesize
  \adjustbox{width=\linewidth}{
  \begin{tabular}{ @{}llll@{} }
  \toprule
      & Batch Size & Learning Rate & Weight Decay \\
    \hline
    LSTM & 32& $1\times10^{-4}$ & $1\times10^{-5}$ \\
    Transformer  & 32 &$1\times10^{-5}$ & $1\times10^{-5}$\\
    \bottomrule
  \end{tabular} }
  \caption{ Hyperparameters used for training the models with LSTM and Transformer encoder respectively. }
  \label{tab:hyperparams}
\end{table}

\section{Additional Results}\label{app:additional}
\subsection{Adversarial Experiments}
\begin{table*}
\ra{1.2}
  \centering
  \setlength\tabcolsep{4.3pt} % default value: 6pt
  \footnotesize
  \adjustbox{width=0.8\textwidth}{
  \begin{tabular}{ @{}lllllll@{} }
  \toprule
      & \multicolumn{2}{c}{\bf IMDb} &  
      \multicolumn{2}{c}{\bf SST} &
      \multicolumn{2}{c}{\bf 20News}\\
      & JSD 
      & Acc.
      & JSD
      & Acc.
      & JSD
      & Acc.\\
    \hline
    BiLSTM (adv. frozen)         & 0.67  
    & 0.76\dm{.14} & 0.62 & 0.76\dm{.06} & 0.67 & 0.78\dm{.13} \\
    BiLSTM (adv. unfrozen)         & 0.67  
    & 0.90\dm{.00}  & 0.61 & 0.82\dm{.00} & 0.67 & 0.91\dm{.00} \\
    \hline
    Transformer (adv. frozen)         & 0.62  
    & 0.71\dm{.18} & 0.57 & 0.76\dm{.03} & 0.62 & 0.87\dm{.03} \\
    Transformer (adv. unfrozen)         & 0.64  
    & 0.87\dm{.02} & 0.57 & 0.78\dm{.01} & 0.62 & 0.92\dpl{.01} \\
    \bottomrule
  \end{tabular} }
  \caption{ JSD (between original and adversarial attention distributions) and model accuracy on test sets. Colored numbers are differences from baseline. While \newcite{wiegreffe2019attention} present TVD, we present JSD as it is the metric being optimized over. Note that JSD is a lower-bound on TVD and should roughly show the same trends.  }
  \label{tab:adversarial}
\end{table*}

We construct adversarial attention distributions by optimizing for the divergence\footnote{Loss function is same as in in \textsection 4 of \newcite{wiegreffe2019attention}} of the distribution from a baseline model's attention distribution using two methods: (1) by transferring all model parameters of a pre-trained base model and optimizing for divergence (frozen) and (2) training an entirely new model and optimizing for divergence (unfrozen). We use Jensen-Shannon divergence (JSD) to measure the difference between the adversarial and baseline distributions.
\cref{tab:adversarial} shows that although we can attain high JSD under both methods, the former leads to a large drop in performance. If we believe the model must encode the same information to achieve similar accuracy, the difference in accuracies of the two methods implies that in the first method, the model likely redistributes information across encoder outputs. 

\subsection{Correlation between Attention Distribution and Inputs/Intermediate Representations}
\begin{figure}[h]
\centering
    \includegraphics[width=\linewidth]{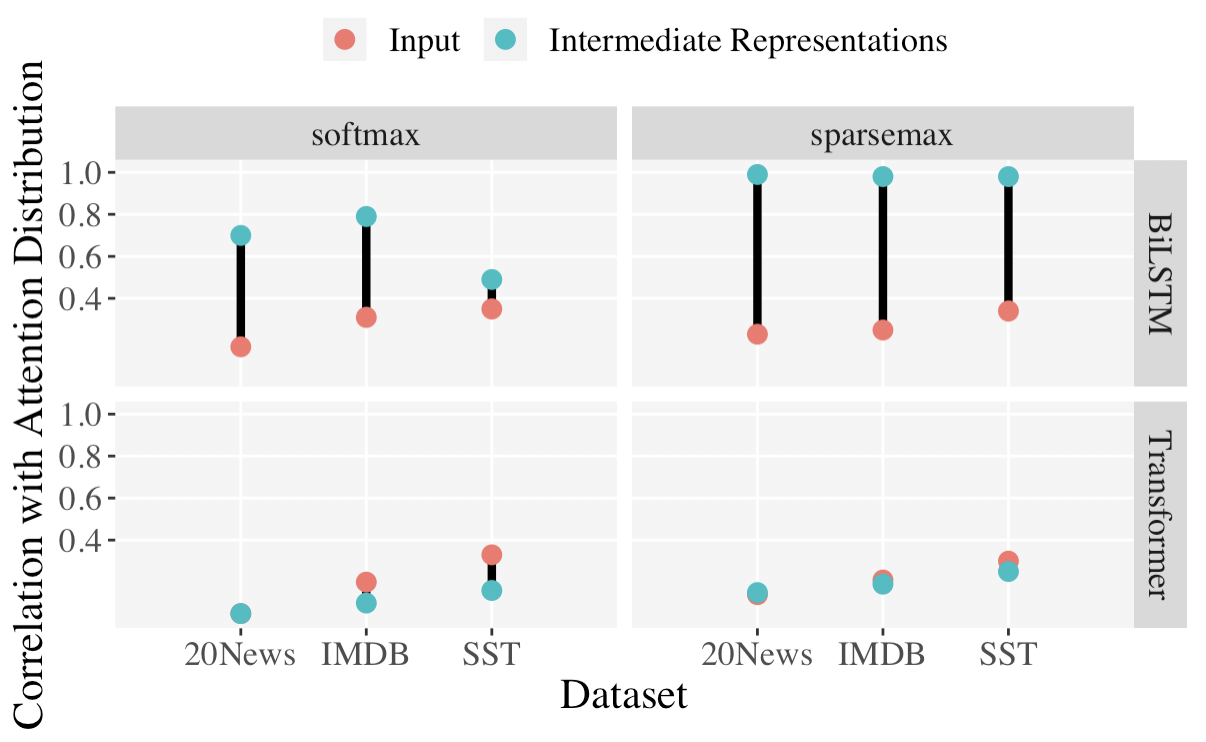}
  \caption{Correlation between the attention distribution and Leave-One_Out FI measures. We see a stronger correlation between attention and intermediate representation FI than input FI across all models.}
    \label{fig:correlation2}
\end{figure}
We provide the full results of our experiments on correlation of the input and intermediate representations with the attention distribution in \cref{tab:input_correlation}. 
\begin{table*}
\ra{1.2}
  \centering
  \setlength\tabcolsep{4.3pt} % default value: 6pt
  \footnotesize
  \adjustbox{width=\textwidth}{
  \begin{tabular}{ @{}lllllllll@{} }
  \toprule
      & \multicolumn{2}{c}{\bf IMDb} & \multicolumn{2}{c}{\bf 20-News} & 
      \multicolumn{2}{c}{\bf SST}\\
      & $\bg_\yhat(\xx)$ & $D_\yhat(\xx)$  & $\bg_\yhat(\xx)$ &  $D_\yhat(\xx)$  & $\bg_\yhat(\xx)$ &  $D_\yhat(\xx)$   \\
    \hline
    BiLSTM Softmax         & 0.36 \dd{0.12}  & 0.31 \dd{0.08}        & 0.24 \dd{0.24} & 0.17 \dd{0.17}& 0.25 \dd{0.28}  & 0.35 \dd{0.18} \\
    BiLSTM Sparsemax         & 0.31 \dd{0.08}  & 0.25 \dd{0.06}        & 0.25 \dd{0.13} & 0.23 \dd{0.09}& 0.45\dd{0.13}  & 0.34 \dd{0.15} \\
    \hline
    Transformer Softmax         & 0.67 \dd{0.08}  & 0.20 \dd{0.10}        & 0.54 \dd{0.11} & 0.05 \dd{0.10}& 0.72 \dd{0.11}  & 0.33 \dd{0.20} \\
    Transformer Sparsemax         & 0.44 \dd{0.08}  & 0.21 \dd{0.09}        & 0.47 \dd{0.10} & 0.14 \dd{0.15}& 0.67 \dd{0.12}  & 0.30 \dd{0.23} \\
    \bottomrule
     & $\bg_\yhat(I)$ & $D_\yhat(I)$  & $\bg_\yhat(I)$ &  $D_\yhat(I)$  & $\bg_\yhat(I)$ &  $D_\yhat(I)$   \rule{0pt}{3ex} \\
    \hline
    BiLSTM Softmax         & 0.77 \dd{0.05}  & 0.79 \dd{0.05}        & 0.78 \dd{0.15} & 0.70 \dd{0.19}& 0.80 \dd{0.13}  & 0.49 \dd{0.17} \\
    BiLSTM Sparsemax         & 0.98 \dd{0.02}  & 0.98 \dd{0.02}        & 0.98 \dd{0.06} & 0.99 \dd{0.01}& 0.97\dd{0.06}  & 0.98 \dd{0.04} \\
    \hline
    Transformer Softmax         & 0.76 \dd{0.04}  & 0.1 \dd{0.07}        & 0.62 \dd{0.13} & 0.05 \dd{0.11}& 0.89 \dd{0.07}  & 0.16 \dd{0.18} \\
    Transformer Sparsemax         & 0.92 \dd{0.05}  & 0.19 \dd{0.09}        & 0.76 \dd{0.15} & 0.15 \dd{0.16}& 0.88 \dd{0.11}  & 0.25 \dd{0.23} \\
    \bottomrule
  \end{tabular} }
  \caption{ Mean Kendall's $\tau$ correlation of attention with gradient-based $\bg_\yhat$ and LOO-based $D_\yhat$ feature importance of the input tokens $\xx$ and intermediate representations $I$. Green numbers are standard deviations. }
  \label{tab:input_correlation}
\end{table*}

%\appendix

\end{document}